\documentclass[conference]{IEEEtran}

\IEEEoverridecommandlockouts

\usepackage{graphicx}
\usepackage{xcolor}

\usepackage{cite}
\usepackage{amsmath}
\usepackage{booktabs}
\usepackage[caption=false]{subfig}
\usepackage{amssymb}
\usepackage{bm}

\usepackage{algorithmic}
\usepackage{algorithm}
\usepackage{comment}

\usepackage[acronym]{glossaries}

\usepackage{amsmath}

\newcommand{\floor}[1]{\left\lfloor #1 \right\rfloor}

\newtheorem{theorem}{\bf Theorem}

\newcommand{\ours}{FGDRA}

\newcommand{\lossFn}{\ell}

\newcommand{\element}{q}
\newcommand{\ELEMENT}{Q}
\newcommand{\elementSet}{\set{\ELEMENT}}

\newcommand{\action}{\phi}
\newcommand{\actionVec}{\vect{\action}}

\newcommand{\actionVecOpt}{\vect{\actionOpt}}

\newcommand{\actionOpt}{\phi\optimal}

\newcommand{\signalTx}{v}
\newcommand{\signalRx}{u}
\newcommand{\noiseTx}{z}

\newcommand{\distance}{d}

\newcommand{\response}{\Omega}
\newcommand{\radiation}{G}
\newcommand{\pathloss}{L}

\newcommand{\distanceTx}{\distance_{\text{RIS},\text{T}}}
\newcommand{\distanceRx}{\distance_{\text{RIS},\text{R}}}

\newcommand{\azimuth}{a}
\newcommand{\azimuthTx}{\azimuth_{\text{RIS},\text{T}}}
\newcommand{\azimuthRx}{\azimuth_{\text{RIS},\text{R}}}

\newcommand{\elevation}{b}
\newcommand{\elevationTx}{\elevation_{\text{RIS},\text{T}}}
\newcommand{\elevationRx}{\elevation_{\text{RIS},\text{R}}}

\newcommand{\scatter}{s}
\newcommand{\SCATTER}{S}

\newcommand{\configuration}{c}
\newcommand{\CONFIGURATION}{C}
\newcommand{\configurationSet}{\set{\CONFIGURATION}}

\newcommand{\rate}{r}

\newcommand{\channelIn}{h}
\newcommand{\channelInVec}{\vect{\channelIn}}
\newcommand{\addLos}[1]{#1_{\text{LOS}}}
\newcommand{\addNlos}[1]{#1_{\text{NLOS}}}

\newcommand{\channelOut}{g}
\newcommand{\channelOutVec}{\vect{\channelOut}}

\newcommand{\mapping}{f}

\newcommand{\inputOrg}{x}
\newcommand{\inputOrgVec}{\vect{\inputOrg}}

\newcommand{\noiseAlone}{ N_0}
\newcommand{\noise}{ \noiseAlone}
\newcommand{\bandwidth}{\omega}

\newcommand{\txpower}{p}

\newcommand{\dataSet}{\set{D}}
\newcommand{\sample}{j}
\newcommand{\SAMPLE}{J}

\newcommand{\worker}{n}
\newcommand{\WORKER}{N}

\newcommand{\modelParam}{\vect{\theta}}

\newcommand{\optMaximize}{\max} 

\newcommand{\expect}{\mathbb{E}\,}

\newcommand{\normalDistribution}{\mathcal{N}}
\newcommand{\normalDistributionComplex}{\mathcal{CN}}

\newcommand{\uniformDistribution}{\mathcal{U}}

\DeclareMathOperator*{\abs}{abs}

\newcommand{\vect}{\boldsymbol}

\newcommand{\seta}[1]{1,\dots,#1}
\newcommand{\set}[1]{\mathcal{#1}}

\newcommand{\one}{\mathbf{1}}
\newcommand{\zero}{\mathbf{0}}

\newcommand{\indict}[2]{\mathbb{I}_{#2}(#1)}

\newcommand{\transpose}{^\dag}
\newcommand{\optimal}{^\star}

\usepackage{forloop}
\newcounter{loopcntr}
\newcommand{\rpt}[2][1]{%
	\forloop{loopcntr}{0}{\value{loopcntr}<#1}{#2}%
}

\newcommand{\subgroup}[1]%
{\rlap{\smash{%
	\newcount\cnt%
	\cnt \numexpr#1\relax%
	\advance\cnt -1\relax%
	$\tabcolsep=.1em\begin{tabular}[t]{|l}\multicolumn{1}{l}{}\\%
	\rpt[\cnt]{\\}
	\\\hline\end{tabular}$%
}}}

\newcounter{myRefCount}

\newacronym{5g}{5G}{fifth generation}

\newacronym{tx}{Tx}{transmitter}

\newacronym{rx}{Rx}{receiver}

\newacronym{ris}{RIS}{reconfigurable intelligent surface}

\newacronym{csi}{CSI}{channel state information}

\newacronym{mle}{MLE}{maximum likelihood estimation}

\newacronym{ml}{ML}{machine learning}

\newacronym{mmw}{mmWave}{millimeter wave}

\newacronym{los}{LOS}{line-of-sight}

\newacronym{nlos}{NLOS}{non line-of-sight}

\newacronym{rss}{RSS}{received signal strength}

\newacronym{snr}{SNR}{signal to noise ratio}

\newacronym{pdf}{PDF}{probability distribution function}

\newacronym{cdf}{CDF}{cummulative density function}

\newacronym{gpr}{GPR}{Gaussian process regression}

\newacronym{nn}{NN}{neural network}

\newacronym{cnn}{CNN}{convolutional neural network}

\newacronym{relu}{ReLU}{rectified linear unit}

\newacronym{erm}{ERM}{empirical risk minimization}

\newacronym{irm}{IRM}{invariant risk minimization}

\newacronym{aoa}{AoA}{angle of arrival}

\newacronym{aod}{AoD}{angle of departure}

\newacronym{mlp}{MLP}{multi layer perceptrons}

\newacronym{ood}{OOD}{out-of distribution}

\newacronym{dro}{DRO}{distributionally robust optimization}

\newacronym{fl}{FL}{federated learning}

\newacronym{ps}{PS}{parameter server}

\allowdisplaybreaks

\begin{document}

\pagenumbering{gobble}
\title{%
	Federated Distributionally Robust Optimization for Phase Configuration of RISs
}

\author{\IEEEauthorblockN{Chaouki Ben Issaid, Sumudu Samarakoon, Mehdi Bennis, and H. Vincent Poor$^\dagger$}
\IEEEauthorblockA{
Centre for Wireless Communications (CWC), University of Oulu, Finland\\
$^\dagger$Electrical Engineering Department, Princeton University, Princeton, USA\\
Email: \{chaouki.benissaid, sumudu.samarakoon, mehdi.bennis\}@oulu.fi, poor@princeton.edu}\ \thanks{This work is supported by Academy of Finland $6$G Flagship (grant no. 318927) and project SMARTER, projects EU-ICT IntellIoT and EUCHISTERA LearningEdge, and CONNECT, Infotech-NOOR, and NEGEIN.}
}



\maketitle
\nopagebreak[4]

\begin{abstract}
In this article, we study the problem of robust \gls{ris}-aided downlink communication over heterogeneous \gls{ris} types in the supervised learning setting. 
By modeling downlink communication over heterogeneous \gls{ris} designs as different workers that learn how to optimize phase configurations in a distributed manner, we solve this distributed learning problem using a distributionally robust formulation in a communication-efficient manner, while establishing its rate of convergence.
By doing so, we ensure that the global model performance of the worst-case worker is close to the performance of other workers.  
Simulation results show that our proposed algorithm requires fewer communication rounds (about 50\% lesser) to achieve the same worst-case distribution test accuracy compared to competitive baselines.
%
%
\end{abstract}
\glsresetall


\section{Introduction}\label{sec:introduction}

Towards enabling \gls{nlos} connectivity, the concept of \glspl{ris} has gained significant interest recently in both industry and academic fora.
Due to the capability of dynamic control of electromagnetic wave propagation using nearly passive multiple reflective elements, \gls{ris} technology is identified as a low-cost and scalable communications solution \cite{He2020,Oezdogan2020}.
%
However, the dynamic configuration of passive reflective elements under changes in the communication system and different \gls{ris} manufacturing designs remains as one of the main challenges in \gls{ris}-aided wireless communication.
The majority of the existing literature on  \glspl{ris}-assisted communication including \cite{Oezdogan2020,Gao2020} and references therein relies on centralized controller-driven optimization and \gls{ml} techniques.
%
Therein, the main focus is to devise \gls{ris} configuration techniques by exploiting statistical correlations within the observed \gls{csi} without distinguishing the impacts of system designs (e.g., differences in propagation environments, \gls{tx}/\gls{rx}/\gls{ris} locations, size of the \gls{ris}, etc.).
%
%
In fact, these works neglect the limitations imposed by communication and privacy concerns during local data sharing, calling for distributed and privacy-preserving approaches. 

\Gls{fl} is a learning framework that allows a centralized model to be trained between several devices and a central entity, a \gls{ps}, while preserving privacy by relying on shared models/gradients rather than accessing their individual data.  
While several federated algorithms have been proposed \cite{kairouz2019, yang2019, li2020}, FedAvg \cite{mcmahan2017} remains the state-of-the-art approach for solving the distributed learning problem in a \gls{ps}-based architecture. In a nutshell, FedAvg is a communication-efficient primal approach that consists of running several local iterations at each worker before exchanging information with the \gls{ps}. 
However, since FedAvg solves the distributed learning problem using the \gls{erm}, i.e. FedAvg minimizes the empirical distribution of the local losses, its performance drops when the local data are non-identically distributed across devices.

The heterogeneity of local data owned by the devices involved in the learning is a significant challenge in \gls{fl} settings compared to classical distributed optimization. In fact, several works \cite{haddadpour2019, karimireddy2020, li2019} have demonstrated that increasing the diversity of local data distributions harms the generalization capability of the central model obtained by solving the distributed learning problem using FedAvg. This is because the \gls{erm} formulation assumes that all local data are drawn from the same distribution. However, this assumption is strong since local data distributions can in practice differ significantly from the average distribution. Consequently, though the global model has a good average performance in terms of test accuracy, its performance locally reduces significantly when the local data are heterogeneous. To obviate this issue, recently, the authors in \cite{Deng2020} proposed a distributionally robust federated averaging (DRFA) algorithm with reduced communication. Instead of using the \gls{erm} formulation, the authors adopt a \gls{dro} objective by formulating the distributed learning problem as a minimization problem of a distributionally robust empirical loss. However, a major weakness with DRFA is that it requires two communication rounds: one to update the primal variable and another one to update the dual variable.

\begin{figure*}
    \centering
    \includegraphics[width=.86\textwidth]{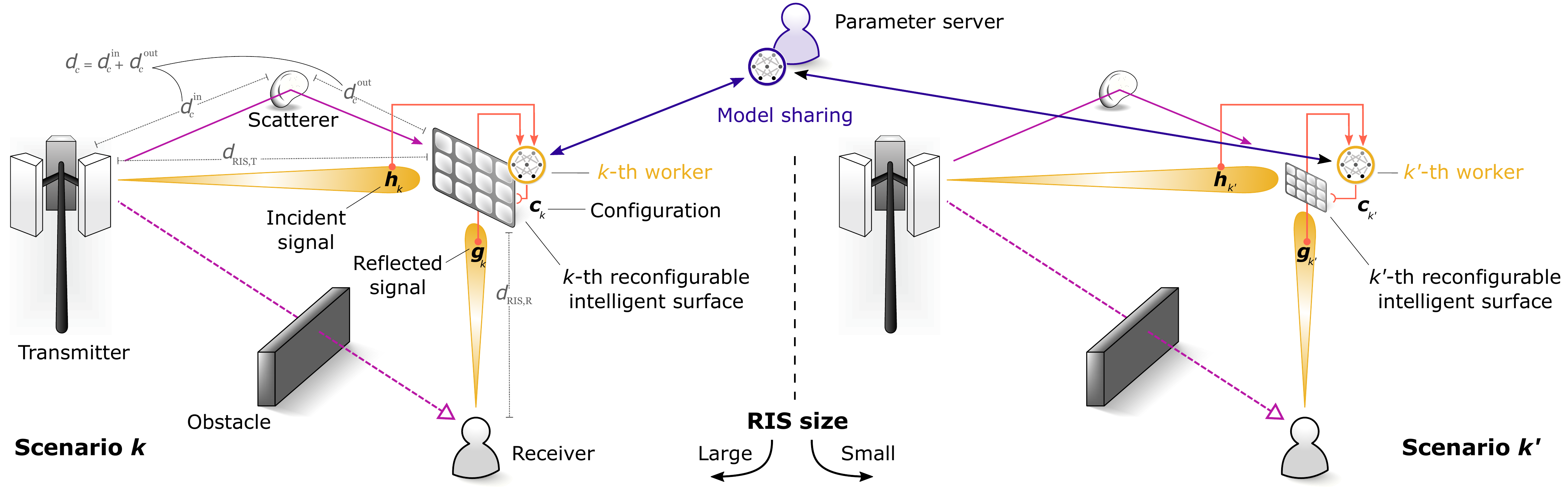}
    \caption{\gls{ris}-aided donwlink communication system highlighting the scenario and worker definitions over different \gls{ris} sizes, as well as the the role of the \gls{ps}.}
    \label{fig:system_model}
\end{figure*}

The main contribution of this paper is to propose a \textbf{communication-efficient and distributionally robust learning algorithm}, dubbed as Federated Group Distributionally Robust Averaging (FGDRA), to learn the optimal \gls{ris} configuration yielding maximum downlink capacity in the heterogeneous network setting. Specifically, we define the dual variables locally for each worker and propose to update them in an adversarial way \cite{sagawa2019, qian2019}, before sharing them with the \gls{ps}, which performs a normalization step to ensure that the dual variables belongs to the simplex. 
As a consequence, the update of the primal and dual variables in our proposed algorithm requires only a single communication round between the devices and the \gls{ps}. Simulation results show that our proposed approach is more communication-efficient than DRFA. Moreover, it incorporates the benefits of the DRO formulation by being more robust to the heterogeneity of local data compared to FedAvg.


\section{System Model \& Problem Formulation}\label{sec:system_model}

We consider a set of multiple downlink \gls{ris} communication scenarios, each consisting of a single \gls{tx}-\gls{rx} pair without \gls{los}, a \gls{ris}, and randomly located scatterers in the \gls{tx}'s vicinity as illustrated in Fig.~\ref{fig:system_model}.
The \gls{nlos} connectivity between \gls{tx} and \gls{rx} is provided by the \gls{ris} via reflecting signals transmitted from \gls{tx} and diffracted signals from a set of $\SCATTER$ scatterers. 
Here, the phases of the \gls{ris} elements are adjusted to maximize the \gls{tx}-\gls{rx} communication data rate by a built-in controller. 
Note that for each scenario, \glspl{ris} with different design specifications in terms of the size of the surface and the distance between \gls{ris} elements are used.
In this view, we define the scenario-specific \gls{ris} controller as a \emph{worker}, hereinafter.
%
%

\subsection{Channel Model}

Let $\channelInVec = [\channelIn_{\element}]_{\element\in\elementSet}$ and $\channelOutVec = [\channelOut_{\element}]_{\element\in\elementSet}$ be the channel vectors of the incident (\gls{tx}-\gls{ris}) and reflected (\gls{ris}-\gls{rx}) signals defined over the reflective elements $\elementSet$ in the \gls{ris}.
The channel model is based on the work in \cite{Basar2020}, in which, the link between \gls{tx} and \gls{ris} is composed of \gls{los} channels as well as \gls{nlos} channels due to the presence of scatterers, while the \gls{ris}-\gls{rx} link has \gls{los} connectivity due to their close proximity.
Let $\distance_o$, $\azimuth_o$, and $\elevation_o$ be the distance, azimuth angle, and elevation angle of an object $o\in\{\text{\gls{tx}},\text{\gls{rx}},\text{Scatterers}\}$ with respect to the \gls{ris}.
With a uniformly distributed random phase $\eta\sim\uniformDistribution[0,2\pi]$ and $\imath^2=-1$, under the assumption that the scatterers are only in the vicinity of the \gls{tx}, the \gls{ris}-\gls{rx} channel is modeled as follows
\begin{equation}\label{eqn:channel_ris_rx}
	\channelOutVec = \sqrt{\radiation(\elevationRx) \pathloss(\distanceRx)}
	e^{\imath\eta} \response(\azimuthRx,\elevationRx),
\end{equation}
where $\radiation(\cdot)$, $\pathloss(\cdot)$, and $\response(\cdot)$ are the \gls{ris} element radiation pattern, distance-dependent path loss, and array response, respectively \cite{Basar2020}.
Similar to \eqref{eqn:channel_ris_rx}, the \gls{los} component of the channel between \gls{ris} and \gls{tx} is modeled by
\begin{equation}\label{eqn:channel_ris_tx_los}
	\addLos{\channelInVec} = 
	\sqrt{\radiation(\elevationTx) \pathloss(\distanceTx)}
	e^{\imath\eta} \response(\azimuthTx,\elevationTx),
\end{equation}
as defined in the \gls{5g} channel model \cite{Basar2020}.
The \gls{nlos} links between the \gls{tx} and \gls{ris} are due to the presence of scatterers.
Let $\distance_{\scatter}$, $\azimuth_\scatter$, and $\elevation_\scatter$ be the  traveled-distance of the reflected signal from \gls{tx} to \gls{ris} at scatterer $\scatter$ and the azimuth and elevation angles of scatterer $\scatter$ with respect to the \gls{ris}, respectively.
Then, the \gls{nlos} channel is modeled as follows
\begin{equation}\label{eqn:channel_ris_tx_nlos}
	\addNlos{\channelInVec} = 
	\textstyle
	\frac{1}{\SCATTER} \sum_{\scatter=1}^\SCATTER \vect{\gamma}_\scatter
	\sqrt{ \radiation(\elevation_\scatter) \pathloss(\distance_\scatter)}
	\response(\azimuth_\scatter,\elevation_\scatter),
\end{equation}
where $\gamma_\scatter\sim\normalDistributionComplex(\zero,\one)$ is a scatterer-dependent random path gain.
In this view, the channel between \gls{tx} and \gls{ris} becomes $\channelInVec = \addLos{\channelInVec} + \addNlos{\channelInVec}$.

\subsection{Downlink Rate Maximization}

For a given scenario, the worker $\worker$ adjust the phases of incident signals at the \gls{ris} to improve the downlink data rate.
Let $\actionVec = [\action_{\element}]_{\element\in\elementSet}$ be the phase change decision at the \gls{ris} over its reflective elements with $\abs( \action_{\element} ) = 1$.
Under which, the received signal $\signalRx$ at the \gls{rx} is given by
\begin{equation}\label{eqn:signal_received}
	\signalRx = 
	\textstyle  \channelOutVec\transpose \actionVec {\channelInVec} \signalTx + \noiseTx,
\end{equation}
where $\signalTx$ is the transmit signal with $\expect[\signalTx^2]=\txpower$
and
$\noiseTx\sim\normalDistribution(0,\noiseAlone)$ is the noise.
The data rate at the \gls{rx} is 
$\rate(\actionVec,\channelInVec,\channelOutVec) = \bandwidth \log_2 \left( 1 + 
\textstyle \frac{| \channelOutVec\transpose \actionVec {\channelInVec} |^2 \txpower }{\bandwidth\noise} \right)$
where $\bandwidth$ is the bandwidth, in which, the downlink data rate maximization is cast as follows
\begin{align}
	\label{eqn:maximize_datarate}
	\underset{\actionVec\in\configurationSet}{\optMaximize} ~
	\rate(\actionVec,\channelInVec,\channelOutVec) = 
	\bandwidth \log_2 \left( 1 + 
	\textstyle \frac{|  \channelOutVec\transpose \actionVec {\channelInVec} |^2 \txpower }{\bandwidth\noise} \right),
\end{align}
where $\configurationSet$ is the feasible set of \gls{ris}  configurations, which are referred to as configuration \emph{classes}.
Due to the notion of configuration classes, an analytical solution cannot be directly derived to determine the optimal configuration $\actionVecOpt$.
The alternate approach is to adopt a heuristic searching mechanism, but the complexity of such a heuristic search increases with the number of reflective elements and their configurations.
Hence, we resort to \gls{ml} to develop a data-driven solution.
Consider that worker $\worker$ has a dataset $\dataSet_\worker = \{ (\inputOrgVec_\sample,\configuration\optimal_\sample) | \sample \in \{\seta{\SAMPLE_\worker}\} \}$ consisting of observed \gls{csi} $\inputOrgVec_\sample = (\channelInVec_\sample, \channelOutVec_\sample)$ and a label $\configuration\optimal_\sample$ corresponds to the optimal \gls{ris} configuration $\actionVecOpt_\sample$.
The data-driven design seeks for a parameterized probabilistic classifier ${\vect{\mapping}}_{\modelParam}(\inputOrgVec_\sample) = [{\mapping}^\configuration_{\modelParam}(\inputOrgVec_\sample)]_{\configuration\in\configurationSet}$ that satisfies
\begin{equation}\label{eqn:centralized}
    \min_{\modelParam} ~
    \textstyle
    -\frac{1}{\WORKER}\sum\limits_{\worker=1}^\WORKER \frac{1}{\SAMPLE_\worker}\sum\limits_{ \sample\in\dataSet_\worker} \sum\limits_{\configuration\in\configurationSet} \indict{\configuration_\sample\optimal}{\configuration}
    \log \big( {\mapping}^\configuration_{\modelParam}(\inputOrgVec_\sample) \big),
\end{equation}
where the indicator $\indict{\configuration_\sample\optimal}{\configuration}=1$ only if the configuration $\configuration$ is equivalent to $\configuration\optimal$, and zero, otherwise.
Note that \eqref{eqn:centralized} relies on a centralized training mechanism where workers share their datasets with a centralized server. 
Under the limitations in data sharing due to communication constraints and/or privacy concerns, \eqref{eqn:centralized} is formulated as a distributed learning problem under the ERM formulation as follows
\begin{equation}\label{ERM}
    \underset{\modelParam}{\min}~ 
    \textstyle \frac{1}{N} \sum\limits_{\worker=1}^\WORKER \lossFn_\worker(\modelParam),
\end{equation}
where $\lossFn_\worker(\modelParam) = - \frac{1}{\SAMPLE_\worker}\sum_{ \sample\in\dataSet_\worker} \sum_{\configuration\in\configurationSet} \indict{\configuration_\sample\optimal}{\configuration}
\log \big( {\mapping}^\configuration_{\modelParam}(\inputOrgVec_\sample) \big)$ is the local loss function of the $n^{th}$ worker.
In \eqref{ERM}, the parameter vector $\modelParam$ is referred to as the \emph{global model}, which can be obtained via the FedAvg algorithm \cite{mcmahan2017}.
Note that under the formulation introduced in \eqref{ERM}, it is assumed that the weight associated with each worker participating in the training is the same, i.e., the centralized model is trained to minimize the loss with respect to the uniform distribution over worker datasets. 
In the presence of heterogeneous local data, relying on the above assumption could result in a model that fails to perform well for some workers yielding a non-robust global model.
%
%
An alternative approach to solving \eqref{ERM} is rather to minimize the distributionally robust empirical loss to learn a model with uniformly good performance across all workers.
Next, we  describe the distributed learning problem under the DRO formulation and elaborate on our approach to solve it.

\section{Distributionally Robust Design of \glspl{ris}}\label{sec:DRO}
We start by stating the DRO formulation for the distributed learning problem as
\begin{align}\label{DRO}
    \underset{\bm{\theta}}{\min}~\underset{\bm{\lambda} \in \Lambda}{\max}~  
    \textstyle F(\bm{\lambda}, \bm{\theta}) = \sum\limits_{n=1}^N \lambda_n \ell_n(\bm{\theta}),
\end{align}
where $\bm{\lambda} \in \Lambda \triangleq \{ \bm{\lambda} \in \mathbb{R}_{+}^N: \sum_{n=1}^N \lambda_n = 1 \}$ is the vector of weights associated with each local loss function. Unlike the \gls{erm} formulation that involves only minimizing over a uniform combination of the loss functions, the DRO formulation is a min-max problem over a weighted sum of the loss functions. Solving the learning problem introduced in \eqref{DRO} ensures the good performance of the global model over the worst-case combination of empirical local distributions. 

Our proposed approach to solve \eqref{DRO} is closely related to the DRFA algorithm proposed in \cite{Deng2020} with a subtle difference in which instead of defining the dual variables vector $\bm{\lambda}$ at the PS side, we define locally for each worker $n$ the dual variable $\lambda_n$. By doing so, we avoid communicating twice between the workers and the PS to update the primal and dual variables, and hence the algorithm is more communication-efficient. For each worker $n$, we define the primal variable as  $\bm{\theta}_n$. Let $K$ denote the number of communication rounds between the PS and the workers, $\tau$ the number of local SGD steps for updating the primal variables, and $B$ the size of the mini-batch used to compute the stochastic gradient. In this case, the total number of iterations is $T = K \tau$. Finally, let $\alpha$ and $\gamma$ denote the learning rates used to update the primal and dual variables, respectively. 

At a given communication round $k$, the {\ours} algorithm runs as follows
\begin{enumerate}
    \item PS selects a subset $\mathcal{S}^k$ of size $m$ from the set $[N] \triangleq \{1,\dots,N\}$ of all workers and send $\bm{\theta}^k$ and $\lambda_n^k$ to each worker $n \in \mathcal{S}^k$.
    \item Each worker $n \in \mathcal{S}^k$ runs locally $\tau$ SGD steps from $\bm{\theta}^k$ to update its primal variable $\bm{\theta}_n^{(k+1)\tau}$.
    \item Given $\bm{\theta}_n^{(k+1)\tau}$, each worker $n \in \mathcal{S}^k$ updates its dual variable $\lambda_n^k$ using an exponentiated gradient ascent, then shares both primal and dual variables with the PS.
    \item PS collects primal variables from every worker $n \in \mathcal{S}^k$ and perform model averaging to update the global model, and then normalizes the dual variables vector. 
\end{enumerate}
\begin{algorithm}[!b] 
	\renewcommand{\algorithmicrequire}{\textbf{Inputs:}}
	\renewcommand{\algorithmicensure}{\textbf{Outputs:}}
	\begin{algorithmic}[1]
		\REQUIRE $N$, $\tau$, $K$, $\alpha$ , $\gamma$, $B$, $m$, $\bm{\theta}^{0}$, $\bm{\lambda}^{0}$.\\
		\ENSURE $\bm{\theta}^K$, $\bm{\lambda}^K$. 
         \FOR{ $k = 0$ to $K-1$ } 
        {\STATE {PS \textbf{samples}  $\mathcal{S}^{k} \subset [N]$ according to  $\bm{\lambda}^{k}$ with size of $m$}\\
        \STATE PS \textbf{broadcasts} $\bm{\theta}^{k}$ and $\lambda_n^{k}$  to each worker $n \in
        \mathcal{S}^{k}$}
        \FOR{worker $n \in \mathcal{S}^{k}$ \textbf{parallel}}
        \STATE Worker \textbf{sets} $\bm{\theta}_n^{k \tau} = \bm{\theta}^{k}$
        \FOR{$t = k \tau,\ldots,(k+1)\tau-1 $}
        \STATE Worker \textbf{samples} mini-batch $\xi^{t}_n$ of size $B$
        \STATE Worker \textbf{updates} its primal variable using, 
        \begin{equation}
        \bm{\theta}^{t+1}_n = \bm{\theta}^{t}_n - \alpha \lambda^{k}_n \nabla \ell_n(\bm{\theta}^{t}_n;\xi^{t}_n)  
        \end{equation}
        \vspace{-20pt}
        \ENDFOR
        \STATE Worker \textbf{updates} its dual variable using, 
        \begin{equation}
            \lambda^{k}_n = \lambda^{k}_n \exp\big(\gamma \ell_n(\bm{\theta}^{(k+1) \tau}_n;\xi_n)\big)
        \end{equation}
        \vspace{-20pt}
        \ENDFOR
        \STATE {Worker $n \in \mathcal{S}^{k}$ \textbf{sends} $\bm{\theta}^{(k+1)\tau}_n$ and $\lambda^{k}_n$ back to the PS}
        \STATE {PS \textbf{computes} $\bm{\theta}^{k+1} = \frac{1}{m} \sum_{n\in \mathcal{S}^{k}}  \bm{\theta}^{(k+1)\tau}_n$ }
        \STATE PS \textbf{normalizes} the dual variables vector $\bm{\lambda}$
        \ENDFOR 
	\end{algorithmic}  
	\caption{Federated Group Distributionally Robust Averaging (FGDRA)}
	\label{alg:1}
\end{algorithm}
\begin{figure*} 
    \centering
  \subfloat[\label{fig1:a}]{%
       \includegraphics[width=0.33\linewidth]{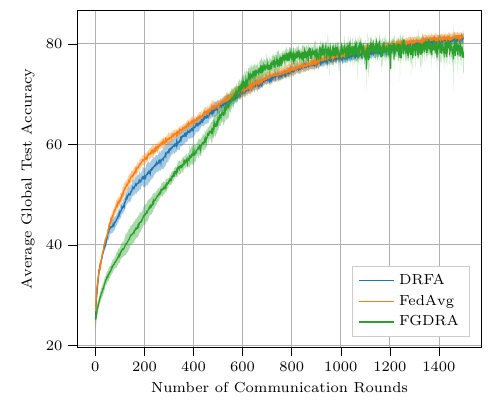}}
    \hfill
  \subfloat[\label{fig1:b}]{%
        \includegraphics[width=0.33\linewidth]{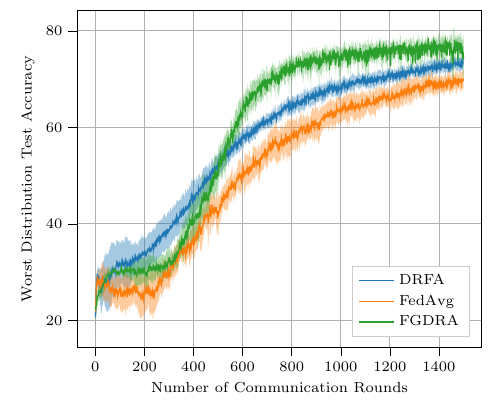}}
    \hfill
  \subfloat[\label{fig1:c}]{%
        \includegraphics[width=0.33\linewidth]{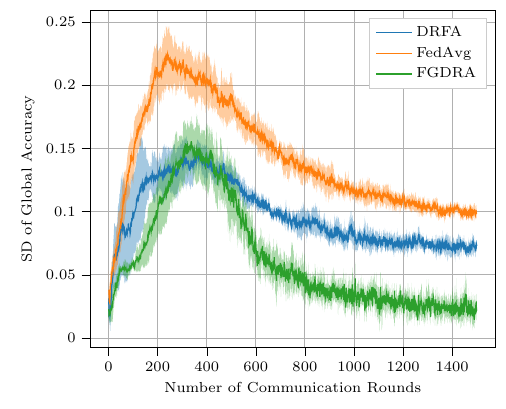}}
  \caption{Comparing {\ours} with DRFA and FedAvg in terms of: (a) average global test accuracy, (b) worst distribution test accuracy, and (c) standard deviation (SD) of the global accuracy.}
  \label{fig1} 
\end{figure*}

Note that while in FedAvg, the PS selects a subset of workers $\mathcal{S}$ randomly in a uniform manner, in our setting (similar to DRFA), the PS selects the subset according to the weighting vector $\bm{\lambda}$. The detailed steps of the {\ours} algorithm are summarized in Algorithm \ref{alg:1}.

Next, we present the theoretical guarantees of our proposed algorithm. First, we state some standard assumptions needed for the proof \\
\textbf{Assumption 1. (Smoothness)} Each local loss function $\ell_n(\cdot), n \in [N]$ and the global function $F(\cdot,\cdot)$ are $L$-smooth.\\
\textbf{Assumption 2. (Bounded Gradient)} There exits a
constant $\sigma > 0$ such that $\mathbb{E}\left[\|\nabla \ell_n(\bm{\theta}, \xi_n)\|\right] \leq \sigma,~\forall n \in [N]$.\\
\textbf{Assumption 3. (Bounded Variance)} There exits a
constant $\nu > 0$ such that $\mathbb{E}\left[\|\nabla \ell_n(\bm{\theta}, \xi_n)-\nabla \ell_n(\bm{\theta})\|\right] \leq \nu,~\forall n \in [N]$.\\
The following theorem establishes the convergence rate of our proposed algorithm.
\begin{theorem}\label{thm}
Suppose assumptions 1-3 hold. If we set $\alpha = \frac{1}{L \sqrt{T}}$, $\gamma = \frac{1}{\sqrt{N}T}$, and $\tau = T^{\frac{1}{4}}$, then we have
\begin{align}
\nonumber & \frac{1}{T} \sum_{t=0}^{T} \mathbb{E}\left[\|\nabla  F\left(\bm{\lambda}^{(\floor{t/\tau})}, \bar{\bm{\theta}}^{t} \right)\|^2 \right]\\
& \leq \left(2 \mathbb{E}\left[F\left(\bm{\lambda}^{0}, \bar{\bm{\theta}}^{0} \right)\right] + \left(\frac{17}{2}+\frac{8}{m}\right) \sigma^2  + 17 \nu^2 \right)\frac{1}{\sqrt{T}}.
\end{align}
\end{theorem}
The proof is deferred to Appendix \ref{appendix}.



\section{Simulation Results}\label{sec:results}
\subsection{Simulation Settings}
In our experiments,  we consider that each worker $n$ has its dataset $\mathcal{D}_n$ generated as detailed in Section II. B. We report both the average global test accuracy, and the worst distribution test accuracy and their corresponding one standard error shaded area based on five runs. The worst distribution test accuracy is defined as the worst of all test accuracies for each local distribution. For fair comparison between the algorithms, we use the same hyperparameters, detailed in Table \ref{table:params1}, unless otherwise stated in the text. We use a multi-layer perceptron (MLP) neural network with two hidden layers having 64 and 32 neurons, respectively, while the input layer and the output layers have 400 and 4 neurons, respectively. The activation function used in the hidden layers is the rectified linear unit (ReLU), while the softmax activation function is used at the output layer. The loss function used is the cross-entropy loss. 
\begin{table}[b]
\caption{Parameters used in the numerical experiments.}
\label{table:params1}
\begin{center}
\begin{small}
\begin{sc}
\begin{tabular}{lc}
\toprule
Parameter & Value \\
\midrule
Learning rate for primal update ($\alpha$)  & $2\times10^{-3}$ \\
Learning rate for dual update ($\gamma$)  & $5\times10^{-3}$ \\
Mini-batch size ($B$) & $50$ \\
Number of workers ($N$)   & $4$ \\
Number of local iterations ($\tau$) & $10$   \\
sampling size ($m$) & $3$ \\
\bottomrule
\end{tabular}
\end{sc}
\end{small}
\end{center}
\vskip -0.25in
\end{table}
\subsection{Communication Efficiency}
We compare the performance of our proposed approach to two baselines, namely DRFA and FedAvg. In Fig. \ref{fig1}, we plot the average global test accuracy, the worst distribution test accuracy, as well as the standard deviation (SD) of the global test accuracy, as a function of the number of communication rounds. We can observe from Fig. \ref{fig1:a} that the performance of the three algorithms in terms of the average global test accuracy is quite similar. However, Fig. \ref{fig1:b} shows that {\ours} outperforms the baselines in terms of the number of communication rounds to achieve the same level of worst distribution test accuracy. By examining Figs. \ref{fig1:a} and \ref{fig1:b} together, we can clearly see that the heterogeneity of the local data has an effect on the performance of the global model. However, the drop in performance is more evident in FedAvg and DRFA compared to our proposed algorithm. Moreover, our approach provides gains in terms of the number of communication rounds compared to DRFA. In fact, {\ours} requires around $800$ communication rounds to converge compared to DRFA requiring more than $1500$ communication rounds. To further support our claim, Fig. \ref{fig1:c} depicts the SD of different workers' accuracy, indicating the degree of fairness of the global model across workers. Compared to FedAvg and DRFA, we clearly see that our proposed approach promotes more fairness among workers in the sense that the global model performs well on the worst distribution compared to the average one.
\subsection{Sensitivity to Hyperparameters}
Next, we study the impact of the number of local iterations $\tau$, the size of the mini-batch $B$ as well as the sampling size $m$ on the average global test accuracy and worst distribution test accuracy, for $K=800$ communication rounds in Table \ref{table:params2}, \ref{table:params3}, and \ref{table:params4} respectively. Table \ref{table:params2} shows that increasing $\tau$ corresponds to an increase in both test accuracies for all algorithms. A similar conclusion can be drawn from Table \ref{table:params3} when increasing $B$, though the impact of $\tau$ seems more noticeable. We report the test accuracies for different values of $m$ in Table \ref{table:params4}. Note that considering a subset of workers participating in the training at each communication round mimics the asynchronous setting. We can observe that increasing $m$ improves both test accuracies for all three algorithms. However, the gap between the average global accuracy and the worst distribution test accuracy is larger in the case of FedAvg and DRFA compared to our proposed approach. If we consider $m=2$, i.e. only half of the workers are sampled, the difference between the average global accuracy and the worst distribution accuracy is $17\%$ in the FedAvg case while it is around $10\%$ for DRFA and about $3\%$ for FGDRA.


\begin{table}[t]
\caption{Impact of the number of local iterations $T$ on the test accuracies\textsuperscript{1}.} %
\label{table:params2}
\begin{center}
\begin{small}
\begin{sc}
\begin{tabular}{lccc}
\toprule
 & $\tau=1$ & $\tau=5$ & $\tau=10$ \\
\midrule
{\ours}  & $54.96/34.12$ & $72.78/65$ & $77.58/73.62$ \\
DRFA  & $46.75/31.62$ & $66.25/44$ & $75.59/64.26$\\
FedAvg & $48.56/24.5$ & $64.7/36.5$ & $73.96/57.62$ \\
\bottomrule
\end{tabular}
\end{sc}
\end{small}
\end{center}
\vskip -0.25in
\end{table}

\begin{table}[t]
\caption{Impact of the mini-batch size $B$ on the test accuracies.}
\label{table:params3}
\begin{center}
\begin{small}
\begin{sc}
\begin{tabular}{lccc}
\toprule
 & $B=10$ & $B=30$ & $B=50$ \\
\midrule
{\ours}  & $70.09/64.87$ & $73.96/69.75$ & $77.58/73.62$ \\
DRFA  & $70.46/58.62$ & $75.28/62$ & $75.59/64.26$\\
FedAvg & $70.28/51.37$ & $76.18/62.37$ & $73.96/57.62$ \\
\bottomrule
\end{tabular}
\end{sc}
\end{small}
\end{center}
\vskip -0.25in
\end{table}

\begin{table}[t]
\caption{Impact of the sampling size $m$ on the test accuracies.}
\label{table:params4}
\begin{center}
\begin{small}
\begin{sc}
\begin{tabular}{lccc}
\toprule
 & $m=1$ & $m=2$ & $m=3$ \\
\midrule
{\ours}  & $57.37/38.5$ & $62.56/59.15$ & $77.58/73.62$ \\
DRFA  & $56.24/36$ & $61.2/51.87$ & $75.59/64.26$\\
FedAvg & $58.06/34.87$ & $62.62/45.12$ & $73.96/57.62$ \\
\bottomrule
\end{tabular}
\end{sc}
\end{small}
\end{center}
\vskip -0.25in
\end{table}
\footnotetext[1]{Test accuracies (expressed in $\%$) are reported in the form (average global test accuracy/worst distribution test accuracy) based on five runs for $K=800$.}

\section{Conclusions and Future Work}\label{sec:conclusion}
%
This work proposes a novel distributed robust \gls{ris}-aided communication design for heterogeneous \gls{ris} configurations. 
The problem is cast as a classification-type \gls{dro} problem
as opposed to data heterogeneity-unaware \gls{erm} approach. To solve this problem, we propose a
communication-efficient and distributionally robust algorithm with convergence guarantees.
As a solution, a neural network-based classifier for phase configuration is trained in a distributed supervised learning manner and compared with two state-of-the-art techniques. 
The results indicate that the proposed classifier is more robust across heterogeneous system designs with faster convergence compared the existing designs.
Future extensions will be focused on systems consisting of multiple transmitters, receivers, and \glspl{ris} with several antennas.  
%

%
\appendices
\section{Proof of Theorem \ref{thm}}\label{appendix}
Let $0 \leq k < K$. For $k \tau \leq t < (k+1) \tau$, we define
\begin{align}
    \textstyle \bar{\bm{\theta}}^t = \frac{1}{m} \sum_{n \in \mathcal{S}^{(\floor{t/\tau})}} \bm{\theta}_n^t.
\end{align}
From the update rule, we have
\begin{align}\label{upd}
    \bm{\theta}_n^{t+1} = \bm{\theta}_n^t - \alpha \lambda_n^k G_n^t,
\end{align}
where $G_n^t = \nabla \ell_n(\bm{\theta}_n^t, \xi_n^t)$. Hence, we can write
\begin{align}
    \bm{\theta}_n^{t+1} 
    &= \textstyle \bm{\theta}_n^{k \tau} - \alpha \lambda_n^k \sum_{r=k \tau}^{t} G_n^r \\
    \bar{\bm{\theta}}^{t+1} 
    &= \textstyle \bm{\theta}_n^{k \tau} -\frac{\alpha}{m} \sum_{j \in \mathcal{S}^{(\floor{t/\tau})}} \sum_{r=k \tau}^{t} \lambda_j^k G_j^t.
\end{align}
For ease of notation, we set $\mathcal{S}_t =\mathcal{S}^{(\floor{t/\tau})}$. Therefore, we have
\begin{align}
    \nonumber& \|\bar{\bm{\theta}}^{t+1} - \bm{\theta}_n^{t+1}\|^2\\
    \nonumber& = \textstyle \left\| \alpha \lambda_n^k \sum_{r=k \tau}^{t} G_n^r - \frac{\alpha}{m} \sum_{j \in \mathcal{S}_t} \lambda_j^k \sum_{r=k \tau}^{t}  G_j^r\right\|^2 \\
    &\leq \textstyle \alpha^2 \tau \sum_{r=k \tau}^{t} \left\|\lambda_n^k G_n^r - \frac{1}{m} \sum_{j \in \mathcal{S}_t} \lambda_j^k G_j^r\right\|^2
\end{align}
%
Adding $\lambda_n^k \nabla \ell_n(\bm{\theta}_n^r) - \lambda_n^k \nabla \ell_n(\bm{\theta}_n^r)$ and $\frac{1}{m} \sum_{j \in \mathcal{S}_t} \lambda_j^k \nabla \ell_j(\bm{\theta}_j^r) - \frac{1}{m} \sum_{j \in \mathcal{S}_t} \lambda_j^k \nabla \ell_j(\bm{\theta}_j^r)$ and using 
%
Assumptions 1 and 2, we get
\begin{align}\label{lem}
    \nonumber& \textstyle \|\bar{\bm{\theta}}^{t+1} - \bm{\theta}_n^{t+1}\|\\
    \nonumber&\leq  \textstyle 4 \alpha^2 \tau (\lambda_{\max}^k)^2 \sum\limits_{r=k \tau}^{t-1} \bigg( \bigg\|\frac{1}{m} \sum\limits_{j \in \mathcal{S}_t} \left(\nabla \ell_j(\bm{\theta}_j^r)-G_j^r\right) \bigg\|^2  \\
    \nonumber& \textstyle + \|\nabla \ell_n(\bm{\theta}_n^r)\|^2 + \|G_n^r - \!\! \nabla \ell_n(\bm{\theta}_n^r)\|^2   + \bigg\|\frac{1}{m} \sum\limits_{j \in \mathcal{S}_t} \!\! \nabla \ell_j(\bm{\theta}_j^r)\bigg\|^2 \bigg) \\
    &\leq \textstyle 4 \alpha^2 \tau^2 \left(\left(1+\frac{1}{m}\right) \sigma^2 + 2 \nu^2 \right),  \!\!\!\!\!\!\!\!\!\!\!\!\!\!\!\!\!\!\!
\end{align}
where $\lambda_{\max}^k = \underset{i \in \mathcal{S}_t}{\max}~\lambda_{i}^k$ and we used  $\lambda_{\max}^k \in (0,1]$.
Using the smoothness of $F$, we have
\begin{align}\label{7}
    \nonumber &\mathbb{E}\left[F\left(\bm{\lambda}^k, \bar{\bm{\theta}}^{t+1} \right)\right] \leq \mathbb{E}\left[F\left(\bm{\lambda}^k, \bar{\bm{\theta}}^{t} \right)\right]\\
    & + \underbrace{\mathbb{E}\left[\langle \nabla F\left(\bm{\lambda}^k, \bar{\bm{\theta}}^{t} \right), \bar{\bm{\theta}}^{t+1}-\bar{\bm{\theta}}^{t} \rangle\right]}_{\text{(I)}} + \frac{L}{2} \underbrace{\mathbb{E}\left[\|\bar{\bm{\theta}}^{t+1}-\bar{\bm{\theta}}^{t}\|^2\right]}_{\text{(II)}}.
\end{align}
From \eqref{upd}, we can write
\begin{equation}
   \bar{\bm{\theta}}^{t+1} = \textstyle \bar{\bm{\theta}}^{t} - \frac{\alpha}{m} \sum_{n \in \mathcal{S}_t} \lambda_n^k G_n^t.
\end{equation}
We start by re-writing the term (II) as
\begin{multline}\label{5}
    \mathbb{E}\left[\|\bar{\bm{\theta}}^{t+1}-\bar{\bm{\theta}}^{t}\|^2\right]
    = \textstyle  \alpha^2 \mathbb{E}\left[\left\|\frac{1}{m} \sum_{n \in \mathcal{S}_t} \lambda_n^k \nabla \ell_n(\bm{\theta}_n^t)\right\|^2\right] \\
    + \textstyle \alpha^2 \mathbb{E}\left[\left\|\frac{1}{m} \sum_{n \in \mathcal{S}_t} \left(\lambda_n^k G_n^t - \lambda_n^k \nabla \ell_n(\bm{\theta}_n^t)\right)\right\|^2\right].
\end{multline}
Focusing on the first term of \eqref{5}, we can write
\begin{multline}\label{6}
    \textstyle \mathbb{E}\left[\left\|\frac{1}{m} \sum_{n \in \mathcal{S}_t} \left(\lambda_n^k G_n^t - \lambda_n^k \nabla \ell_n(\bm{\theta}_n^t)\right)\right\|^2\right]\\
    \textstyle \leq \frac{(\lambda_{\max}^k)^2}{m^2} \sum_{n \in \mathcal{S}_t} \mathbb{E}\left[\left\| G_n^t - \nabla \ell_n(\bm{\theta}_n^t)\right\|^2\right] \leq \frac{\sigma^2}{m},
\end{multline}
where we used that $\lambda_{\max}^k \in (0,1]$ and Assumption 3. Replacing \eqref{6} in \eqref{5}, we get
\begin{equation}\label{10}
\textstyle \mathbb{E}\left[\|\bar{\bm{\theta}}^{t+1}-\bar{\bm{\theta}}^{t}\|^2\right] \leq \frac{\alpha^2 \sigma^2}{m} + \alpha^2 \mathbb{E}\bigg[\Big\| \sum\limits_{n \in \mathcal{S}_t} \frac{\lambda_n^k}{m} \nabla \ell_n(\bm{\theta}_n^t)\Big\|^2\bigg].
\end{equation}
For term (I), we can write 
\begin{multline}
\mathbb{E}\left[\langle \nabla F\left(\bm{\lambda}^k, \bar{\bm{\theta}}^{t} \right), \bar{\bm{\theta}}^{t+1}-\bar{\bm{\theta}}^{t} \rangle\right]\\
 \textstyle = - \alpha \mathbb{E}\left[\langle \nabla F\left(\bm{\lambda}^k, \bar{\bm{\theta}}^{t} \right), \frac{1}{m} \sum_{n \in \mathcal{S}_t} \lambda_n^k G_n^t\rangle\right]
\end{multline}
Using the unbiaseness of $G_n^t$ and the identity $\langle a, b\rangle = \frac{1}{2}\left(\|a\|^2+\|b\|^2-\|a-b\|^2\right)$, we get
\begin{align}\label{12}
\nonumber & \mathbb{E}\left[\langle \nabla F\left(\bm{\lambda}^k, \bar{\bm{\theta}}^{t} \right), \bar{\bm{\theta}}^{t+1}-\bar{\bm{\theta}}^{t} \rangle\right]\\
\nonumber &= \textstyle -\frac{\alpha}{2} \mathbb{E}\left[\|\nabla F\left(\bm{\lambda}^k, \bar{\bm{\theta}}^{t} \right)\|^2\right] - \frac{\alpha}{2} \mathbb{E}\left[\left\| \sum_{n \in \mathcal{S}_t} \frac{\lambda_n^k}{m} \nabla \ell_n(\bm{\theta}_n^t)\right\|^2\right]\\
&\textstyle + \frac{\alpha}{2} \underbrace{\textstyle \mathbb{E}\left[\left\|\nabla F\left(\bm{\lambda}^k, \bar{\bm{\theta}}^{t} \right)-\frac{1}{m} \sum_{n \in \mathcal{S}_t} \lambda_n^k \nabla \ell_n(\bm{\theta}_n^t)\right\|^2\right]}_{(III)}.
\end{align}
Using \cite[Lemma 1]{Deng2020}, we can re-write the term (III) as
\begin{align}\label{14}
    \nonumber &\textstyle \mathbb{E}\left[\left\|\nabla F\left(\bm{\lambda}^k, \bar{\bm{\theta}}^{t} \right)-\frac{1}{m} \sum_{n \in \mathcal{S}_t} \lambda_n^k \nabla \ell_n(\bm{\theta}_n^t)\right\|^2\right]\\
     \nonumber & \quad \textstyle \leq \mathbb{E}\left[\left\|\frac{1}{N} \sum_{n=1}^N \lambda_n^k \left(\nabla \ell_n\left(\bar{\bm{\theta}}^{t} \right)-\nabla \ell_n\left(\bm{\theta}_n^{t} \right)\right)\right\|^2\right]\\
    \nonumber & \quad\textstyle \leq \frac{2L^2}{N} \mathbb{E}\left[\sum_{n=1}^N  \|\bar{\bm{\theta}}^{t} - \bm{\theta}_n^{t}\|^2\right]\\
    & \quad\textstyle \leq 8 \alpha^2 L^2 \tau^2 \left(\left(1+\frac{1}{m}\right) \sigma^2 + 2 \nu^2\right),
\end{align}
where we used \eqref{lem}. Replacing \eqref{10}, \eqref{12}, and \eqref{14} in \eqref{7}, re-arranging the terms and using $0 \leq \alpha \leq \frac{1}{L}$, we get
\begin{multline}\label{15}
\textstyle \mathbb{E}\left[\|\nabla F\left(\bm{\lambda}^k, \bar{\bm{\theta}}^{t} \right)\|^2 \right] \leq \frac{2 \left(\mathbb{E}\left[F\left(\bm{\lambda}^k, \bar{\bm{\theta}}^{t} \right)\right]-\mathbb{E}\left[F\left(\bm{\lambda}^k, \bar{\bm{\theta}}^{t+1} \right)\right]\right)}{\alpha}\\
\textstyle  + \frac{\sigma^2 \alpha L}{2} + 8 \alpha^2 L^2 \tau^2 \left(\left(1+\frac{1}{m}\right) \sigma^2 + 2 \nu^2\right).
\end{multline}
Next, we decompose the term $\mathbb{E}\left[F\left(\bm{\lambda}^k, \bar{\bm{\theta}}^{t} \right)\right]-\mathbb{E}\left[F\left(\bm{\lambda}^k, \bar{\bm{\theta}}^{t+1} \right)\right]$ by writing 
\begin{multline}\label{16}
    \mathbb{E}\left[F\left(\bm{\lambda}^k, \bar{\bm{\theta}}^{t} \right)\right]-\mathbb{E}\left[F\left(\bm{\lambda}^k, \bar{\bm{\theta}}^{t+1} \right)\right]\\
    = \mathbb{E}\left[F\left(\bm{\lambda}^k, \bar{\bm{\theta}}^{t} \right)\right]-\mathbb{E}\left[F\left(\bm{\lambda}^{k+1}, \bar{\bm{\theta}}^{t+1} \right)\right] \\
    + \mathbb{E}\left[F\left(\bm{\lambda}^{k+1}, \bar{\bm{\theta}}^{t+1} \right)\right]-\mathbb{E}\left[F\left(\bm{\lambda}^k, \bar{\bm{\theta}}^{t+1} \right)\right].
\end{multline}
Let $\bm{\ell}(\bm{\theta}) = [\ell_1(\bm{\theta}), \dots, \ell_N(\bm{\theta})]^T$, then we can write
\begin{align}\label{17}
\nonumber & \mathbb{E}\left[F\left(\bm{\lambda}^{k+1}, \bar{\bm{\theta}}^{t+1} \right)\right]-\mathbb{E}\left[F\left(\bm{\lambda}^{k}, \bar{\bm{\theta}}^{t+1} \right)\right]\\
\nonumber &\qquad \overset{\mathrm{(a)}}{\leq} \mathbb{E}\left[\|\bm{\lambda}^{k+1}-\bm{\lambda}^{k}\|\right] \mathbb{E}\left[\|\bm{\ell}(\bar{\bm{\theta}}^{t+1})\|\right]\\
&\qquad \overset{\mathrm{(b)}}{\leq} \sqrt{N} \nu \mathbb{E}\left[\|\bm{\lambda}^{k+1}-\bm{\lambda}^{k}\|\right] \overset{\mathrm{(c)}}{\leq} \textstyle \frac{\sqrt{N}}{2} \gamma \nu^2, 
\end{align}
where we used Cauchy–Schwarz inequality in $\mathrm{(a)}$, Assumption 2 in $\mathrm{(b)}$, and  Pinsker’s inequality  and \cite[Eq (9)]{qian2019} in $\mathrm{(c)}$. Using \eqref{16} and \eqref{17} in \eqref{15}, we get
\begin{align}
\nonumber & \mathbb{E}\left[\|\nabla F\left(\bm{\lambda}^{(\floor{t/\tau})}, \bar{\bm{\theta}}^{t} \right)\|^2 \right]\\
\nonumber &\textstyle \leq \frac{2 \left(\mathbb{E}\left[F\left(\bm{\lambda}^{(\floor{t/\tau})}, \bar{\bm{\theta}}^{t} \right)\right]-\mathbb{E}\left[F\left(\bm{\lambda}^{(\floor{t/\tau})+1}, \bar{\bm{\theta}}^{t+1} \right)\right]\right)}{\alpha}\\
&\textstyle + \frac{\sigma^2 \alpha L}{2} + \frac{\sqrt{N} \gamma \nu^2}{\alpha} + 8 \alpha^2 L^2 \tau^2 \left(\left(1+\frac{1}{m}\right) \sigma^2 + 2 \nu^2\right).
\end{align}
Equivalently, we can write
\begin{multline}
\textstyle \frac{1}{T} \sum_{t=0}^{T} \mathbb{E}\left[\|\nabla  F\left(\bm{\lambda}^{(\floor{t/\tau})}, \bar{\bm{\theta}}^{t} \right)\|^2 \right]\\
\textstyle \leq \frac{2}{\alpha T} \mathbb{E}\left[F\left(\bm{\lambda}^{0}, \bar{\bm{\theta}}^{0} \right)\right] + \frac{\sigma^2 \alpha L}{2} + \frac{\sqrt{N} \gamma \nu^2}{\alpha} \\
\textstyle   + 8 \alpha^2 L^2 \tau^2 \left(\left(1+\frac{1}{m}\right) \sigma^2 + 2 \nu^2\right).
\end{multline}
Choosing $\alpha = \frac{1}{L \sqrt{T}}$, $\gamma = \frac{1}{\sqrt{N}T}$, and $\tau = T^{\frac{1}{4}}$, we can write
\begin{multline}
\textstyle \frac{1}{T} \sum_{t=0}^{T} \mathbb{E}\left[\|\nabla  F\left(\bm{\lambda}^{(\floor{t/\tau})}, \bar{\bm{\theta}}^{t} \right)\|^2 \right]\\
\textstyle\leq \left(2 \mathbb{E}\left[F\left(\bm{\lambda}^{0}, \bar{\bm{\theta}}^{0} \right)\right] + \left(\frac{17}{2}+\frac{8}{m}\right) \sigma^2  + 17 \nu^2 \right)\frac{1}{\sqrt{T}}.
\end{multline}

\bibliographystyle{IEEEtran}
\bibliography{references}

\end{document}